\documentclass[10pt,twocolumn,letterpaper]{article}

\usepackage[pagenumbers]{cvpr} 

\usepackage{amsmath}
\usepackage{amssymb}
\usepackage{booktabs}
\usepackage{multirow}
\usepackage{gensymb}
\usepackage{makecell}
\usepackage{stfloats}
\usepackage{dsfont}

\definecolor{cvprblue}{rgb}{0.21,0.49,0.74}
\usepackage[pagebackref,breaklinks,colorlinks,allcolors=cvprblue]{hyperref}

\def\paperID{*****} 
\def\confName{CVPR}
\def\confYear{2026}

\title{Joint Instance Segmentation and Geometric Attribute Regression\\for Roof Structures in Aerial Imagery}

\author{Luuk Versteeg$^{1}$\quad
Rob G.J. Wijnhoven$^{2}$\quad
Martin R. Oswald$^{1}$\\
$^{1}$University of Amsterdam (UvA), The Netherlands\quad
$^{2}$Spotr.ai, The Hague, The Netherlands\\
{\tt\small luukversteeg@live.nl\quad rob@spotr.ai\quad m.r.oswald@uva.nl}
}

\begin{document}
\maketitle

\begin{abstract}
We present a method for jointly predicting instance-level roof segment masks together with three continuous geometric attributes---building height, roof slope, and roof azimuth---from a single aerial orthophoto. Our approach extends Mask R-CNN with a dedicated attribute regression branch and introduces two key innovations: a conditional azimuth loss that suppresses supervision for flat roof segments where azimuth labels are inherently noisy, and a log-normalized height representation that addresses the heavily skewed distribution of building heights. We train and evaluate on a large-scale dataset of Dutch aerial images paired with automatically derived ground truth from 3DBAG, a nationwide LiDAR-based 3D building dataset. Using a DINOv3 ConvNeXt-Base backbone, our method achieves a mean absolute error of approximately 4 degrees for roof slope, 7 degrees for azimuth, and 1 meter for building height, with an instance segmentation AP$_{50}$ of 0.566. The predicted per-segment masks and attributes are sufficient to reconstruct simplified 3D building models (LoD2) from a single overhead image, requiring expensive 3D reference data only for training.
\end{abstract}

\begin{figure}[t]
  \centering
  \includegraphics[width=\linewidth]{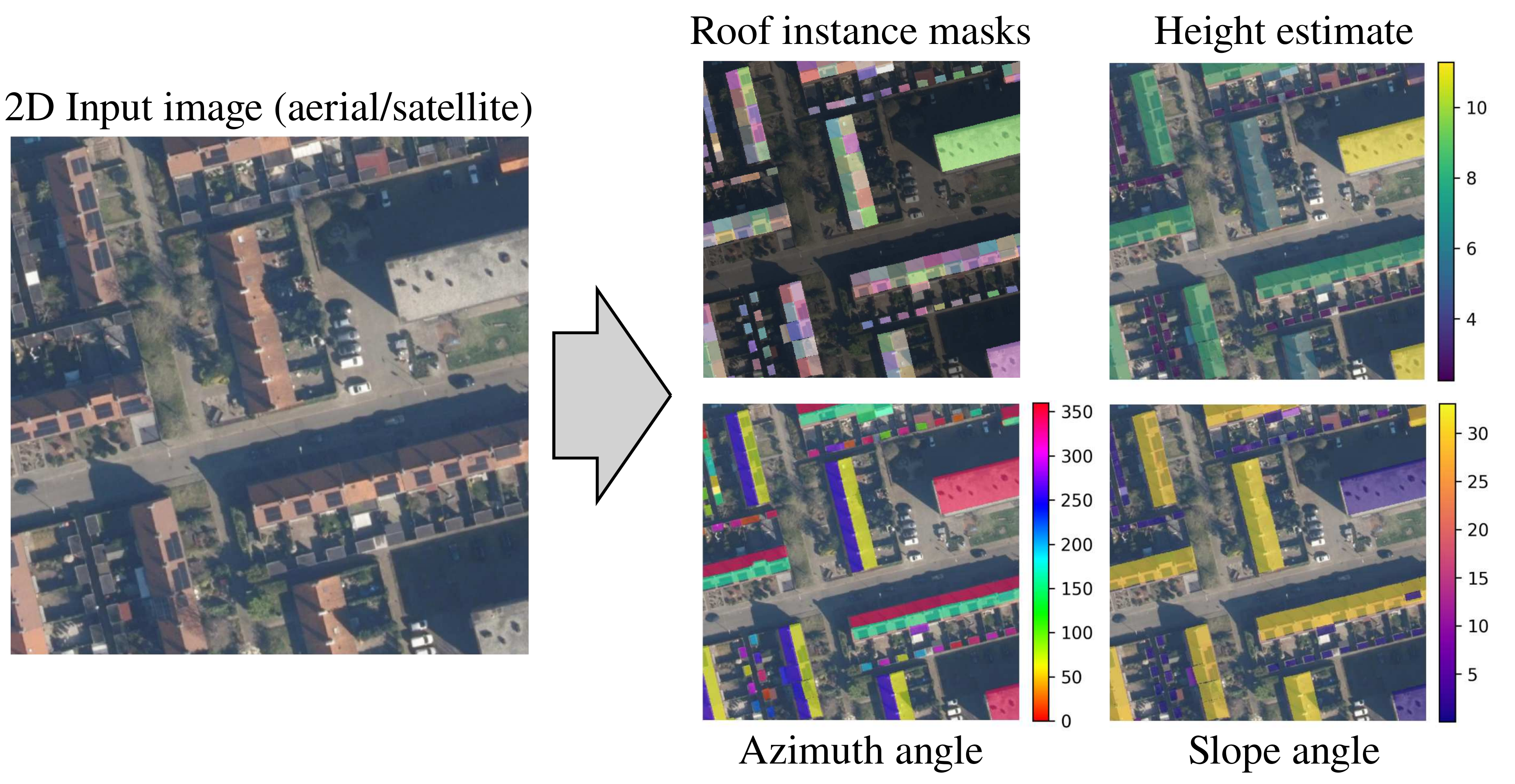}
  \caption{Overview of the proposed method. From a single aerial orthophoto, the model jointly predicts instance-level roof segment masks and three geometric attributes per segment: building height, roof slope (angle), and compass orientation (azimuth).}
  \label{fig:teaser}
\end{figure}
\section{Introduction}
\label{sec:intro}

Buildings account for 40\% of the European Union's energy consumption and 36\% of its greenhouse gas emissions, yet less than 1\% of the building stock is renovated annually~\cite{eu_buildings}. Addressing this at scale---whether through decarbonization programs, energy label compliance, or solar potential assessment---requires detailed knowledge of the built environment. Geometric building attributes such as height, roof slope, and orientation determine solar panel suitability, thermal performance, and building volume. Beyond sustainability, these attributes are critical for the insurance and housing sectors. Rebuild value estimation depends on an accurate assessment of building volume, which in turn requires height, footprint area, and roof shape. Underinsurance remains a widespread problem, in part because accurate geometric data is simply unavailable at the scale of national building stocks. A scalable method for acquiring these attributes would therefore benefit sustainability planning, insurance risk assessment, and housing management alike.

Several data acquisition strategies exist for obtaining building geometry, each with different trade-offs between resolution, coverage, and cost. Manual inspection and drone surveys provide high-quality measurements but require physical site visits, limiting their applicability to targeted inspections rather than national-scale indexing. LiDAR offers accurate 3D measurements, yet dedicated airborne campaigns are expensive and infrequent. The Dutch Actueel Hoogtebestand Nederland (AHN) is among the best national LiDAR programs, collected in 2--3 year cycles, but most countries lack comparable infrastructure. In the Netherlands, AHN data has been processed into 3DBAG~\cite{3dbag}---a nationwide dataset of LoD2 building reconstructions providing per-segment height, roof slope, and azimuth---but such processed 3D building datasets remain exceptional. Multi-view reconstruction from street-level imagery is theoretically appealing, yet it is hampered in practice by sparse views, unreliable camera calibration, and frequent occlusions from vegetation, vehicles, and neighboring structures~\cite{vggt}. Aerial and satellite imagery, by contrast, is widely available globally at high resolution (8--25~cm/pixel), and many countries maintain national orthophoto programs. This makes single-image approaches from overhead data the most scalable path for large-scale building attribute estimation.

Our key insight is that expensive 3D reference data such as 3DBAG can be used as \emph{ground truth for training}, enabling a model that infers 3D geometric attributes---including building height---from a single 2D aerial image at inference time. This means the model can be deployed anywhere aerial imagery is available---the expensive 3D data is needed only once, to train the model.

We make the following contributions:
\begin{enumerate}
    \item We extend Mask R-CNN with an attribute regression branch for joint prediction of instance-level roof segment masks and three continuous geometric attributes: building height, roof slope, and roof azimuth.
    \item We introduce a conditional azimuth loss that gates supervision based on roof slope, suppressing noisy azimuth labels for flat roof segments and nearly halving the azimuth prediction error.
    \item We show that a log-normalized height representation improves height regression by 17\% over direct regression, by reshaping the skewed height distribution into a near-Gaussian target.
    \item We provide a large-scale evaluation on 17,353 aerial images with over 2 million instance annotations, including an analysis of the strengths and limitations of 3DBAG as automatically derived ground truth.
\end{enumerate}

\section{Related Work}
\label{sec:related}

Classical approaches to building geometry estimation rely on Structure-from-Motion, multi-view stereo, or shadow analysis, but require multiple calibrated views or specific acquisition conditions that limit scalability. This has motivated learning-based methods that infer geometry from single images.

\paragraph{Building height estimation from imagery.}
Estimating building height from remote sensing data has received considerable attention. Olson and Saxe~\cite{olson2024} demonstrate that building height can be regressed from single street-view images using EfficientNet, achieving a mean absolute error (MAE) of 1.21~m on Toronto buildings, showing that visual cues alone carry substantial height information. For instance-level prediction, Mask-to-Height~\cite{mask2height2025} combines YOLOv11-based instance segmentation with height estimation from satellite imagery, achieving 60.4\% mAP$_{50}$ on the DFC2023 benchmark---but discretizes height into five classes rather than predicting continuous values. 3DCentripetalNet~\cite{3dcentripetalnet} introduces a 3D centripetal shift representation for estimating building heights from single aerial images, though performance degrades with tilted imagery and complex roof geometries. Most recently, BuildMamba~\cite{buildmamba2026} proposes a state-space model for joint building segmentation and continuous height regression from satellite images, though it operates at the semantic rather than instance level. A common limitation across these methods is their exclusive focus on height, ignoring other geometric attributes such as roof slope and orientation that are essential for full 3D building characterization.

\paragraph{Roof slope and orientation estimation.}
Roof slope and azimuth estimation has primarily been studied for solar potential assessment. SolarMTNet~\cite{solarmtnet2025} simultaneously segments roof orientations and slopes from aerial imagery, achieving a mean IoU of 0.67 for orientation and 0.40 for slope. However, it operates at the semantic segmentation level---predicting per-pixel maps rather than per-instance attributes---which cannot distinguish multiple roof segments with different geometries on a single building. SolarNet~\cite{solarnet2022} similarly targets rooftop orientation for solar potential but does not predict slope.

\paragraph{Instance segmentation in aerial imagery.}
Mask R-CNN~\cite{maskrcnn} remains widely used for building footprint extraction and roof segmentation in remote sensing applications. Mask2Former~\cite{mask2former}, the current state-of-the-art for instance segmentation on standard benchmarks, has been adapted for remote sensing~\cite{iq2former2024}, but its computational requirements can be prohibitive for domain-specific datasets with long training cycles. Most instance segmentation work in remote sensing focuses on detection and mask prediction alone, without predicting geometric attributes per instance.

\paragraph{3D building reconstruction.}
LiDAR-based approaches such as 3DBAG~\cite{3dbag} provide accurate 3D building models at national scale, but require expensive data collection infrastructure. Learning-based multi-view reconstruction methods such as VGGT~\cite{vggt} aim to reconstruct 3D scenes from images, but struggle with sparse and poorly calibrated input views that are typical in practical building inspection settings. Our approach sidesteps full 3D reconstruction: by predicting per-segment masks and geometric attributes from a single aerial image, we obtain sufficient information to reconstruct simplified LoD2 building models without requiring multi-view input or LiDAR at inference time.

\paragraph{Summary.} To our knowledge, no prior work jointly predicts instance-level building masks with continuous regression of height, roof slope, and azimuth from a single aerial image.

\section{Method}
\label{sec:method}

\begin{figure}[t]
  \centering
  \includegraphics[width=\linewidth]{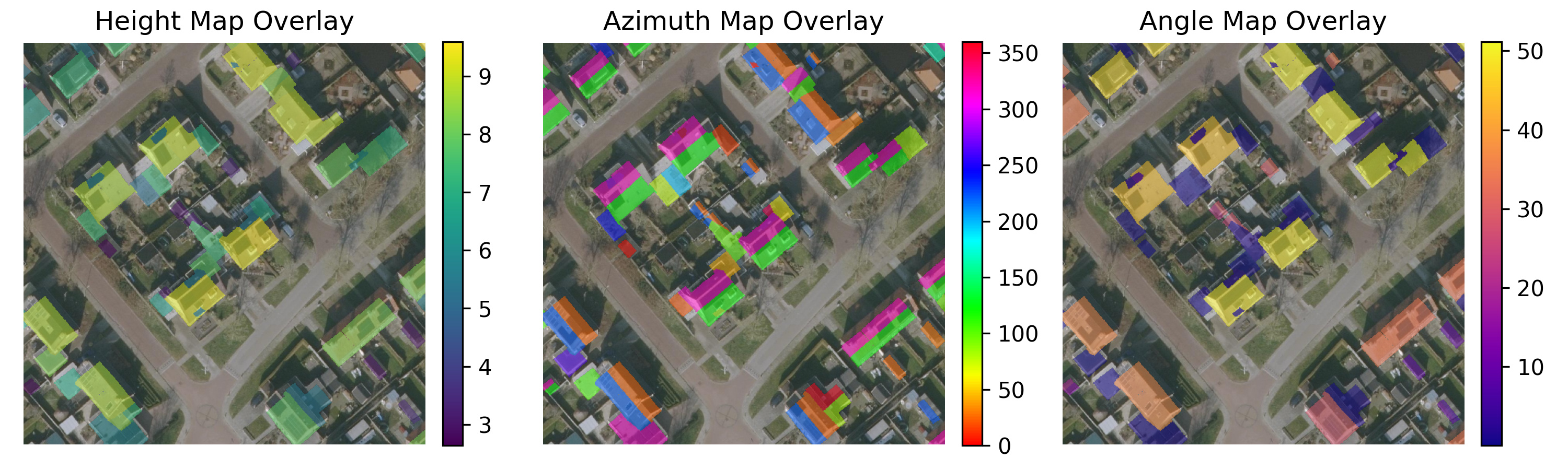}
  \caption{Example from our dataset. Top left: aerial image. Top right: instance segmentation masks (LoD2.2 roof segments). Bottom: per-pixel attribute maps for height (meters), azimuth (degrees), and roof angle (degrees).}
  \label{fig:example}
\end{figure}

We frame the task as instance segmentation with continuous attribute regression at the roof-segment level. For each detected roof segment, the model predicts a binary mask and three geometric attributes: building height, roof slope (angle), and roof azimuth (compass orientation). Together, these outputs provide sufficient information to reconstruct a simplified 3D building model at LoD2 level of detail.

\subsection{Dataset}
\label{sec:dataset}

\paragraph{Aerial imagery.}
We use the PDOK Luchtfoto RGB dataset~\cite{pdok}, a collection of high-resolution aerial orthophotos of the Netherlands captured at 8~cm per pixel resolution. These images are acquired annually during summer flights and are freely available as OGC-compliant web services. We use the final-release orthophotos, which provide geometrically corrected, nadir-view imagery suitable for quantitative analysis.

\paragraph{Ground truth from 3DBAG.}
To obtain geometric ground-truth labels without manual annotation, we leverage 3DBAG~\cite{3dbag}, a nationwide dataset of 3D building models constructed by combining cadastral building footprints (BAG) with airborne LiDAR point clouds (AHN). 3DBAG provides building reconstructions at multiple levels of detail. We use LoD2.2, the most detailed level, which models individual roof surfaces with their height, slope, and orientation rather than representing buildings as simple extruded blocks. Since 3DBAG is derived from BAG polygons and LiDAR measurements, its accuracy is bounded by the quality of these underlying sources, introducing small but systematic uncertainties in height and roof geometry.

\paragraph{Dataset construction.}
Starting from Dutch addresses in an existing production pipeline, we select one building per address and generate a $100 \times 100$~meter square crop centered on the building centroid. The corresponding aerial image is downloaded from PDOK and resized to $1024 \times 1024$~pixels. Using the same spatial extent, we query 3DBAG within a PostgreSQL database to extract all LoD2.2 roof segments and their geometric attributes, which are converted into instance segmentation masks. The resulting dataset comprises 17,353 paired orthophotos and annotation maps containing a total of 2,110,801 instance masks (122 per image on average). We partition the data into training (11,931), validation (2,747), and test (2,675) sets using a 60/15/15 split with strict geographic separation: all buildings within a 1~km radius are clustered together and assigned to the same split, preventing spatial leakage between subsets.

\begin{figure}[t]
  \centering
  \includegraphics[width=\linewidth]{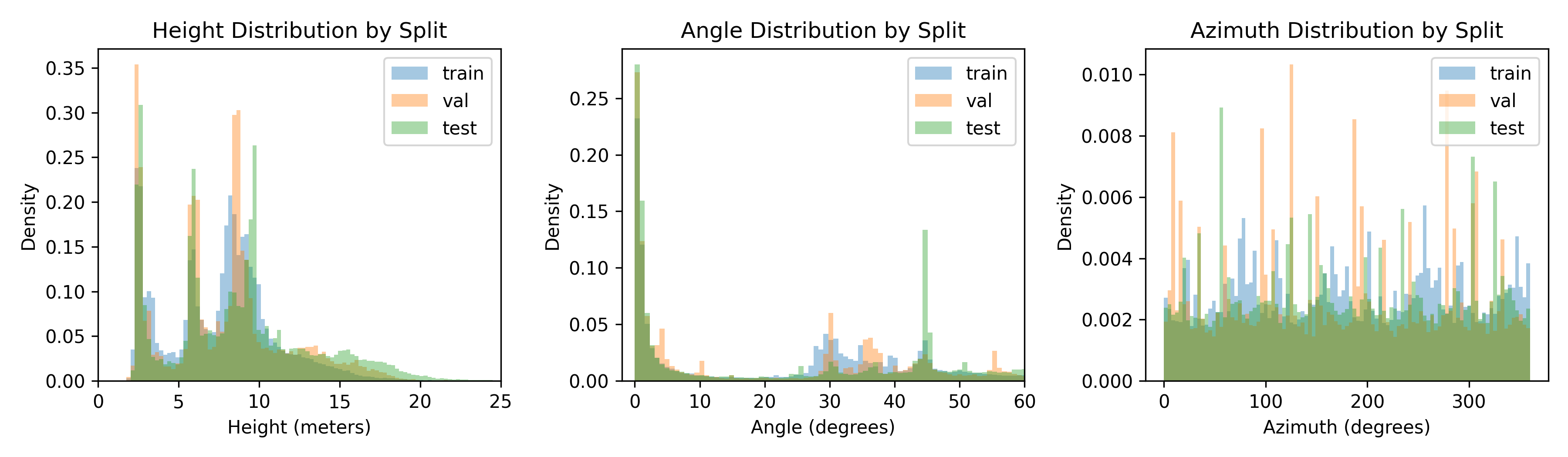}
  \caption{Distribution of building attributes across dataset splits. Height shows three peaks at common Dutch residential building types. Roof angle has a large concentration at $0\degree$ (flat roofs). Azimuth is approximately uniform.}
  \label{fig:distributions}
\end{figure}

\paragraph{Attribute distributions.}
Figure~\ref{fig:distributions} shows the variation per attribute in the dataset. The height distribution is heavily right-skewed, with three dominant peaks at approximately 3, 6, and 9~meters corresponding to common Dutch residential building types, and a long tail extending to over 100~meters for church towers and high-rise buildings. Roof angles show a large concentration near $0\degree$ (flat roofs), with secondary peaks near $30\degree$ and $45\degree$ for sloped roofs. The azimuth distribution is approximately uniform, as roof orientations are not tied to a preferred global direction. Notably, flat roof segments in 3DBAG are assigned arbitrary azimuth values, since a horizontal surface has no meaningful compass orientation---this observation motivates our conditional azimuth loss (Sec.~\ref{sec:loss}).

\subsection{Architecture}
\label{sec:architecture}

Our model is based on Mask R-CNN~\cite{maskrcnn} with a Feature Pyramid Network (FPN)~\cite{fpn}. The standard Mask R-CNN pipeline detects candidate regions via a Region Proposal Network, then predicts per-instance bounding boxes, class labels, and binary masks through separate branches operating on RoI-aligned features. Two MultiScaleRoIAlign modules extract fixed-size feature maps from proposed regions: $7 \times 7$ features for the classification, bounding box, and attribute branches, and $14 \times 14$ features for the mask branch, which requires higher spatial resolution for pixel-level prediction.

\begin{figure}[t]
  \centering
  \includegraphics[width=\linewidth]{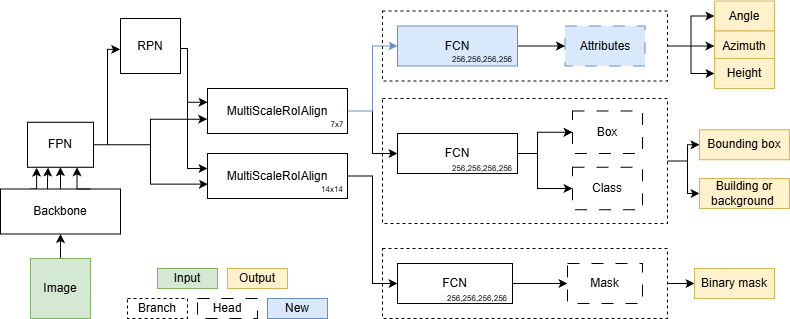}
  \caption{Model architecture. Mask R-CNN is extended with an Attribute Branch (highlighted in blue) that regresses roof slope, azimuth, and building height for each detected roof segment.}
  \label{fig:architecture}
\end{figure}

We extend this architecture with an \textbf{Attribute Branch} that regresses continuous geometric properties for each detected roof segment. This branch shares the $7 \times 7$ RoI-aligned features with the classification and bounding box branch. It consists of a fully convolutional subnetwork (four convolutional layers with Batch Normalization) followed by a small linear head (two layers, 128 hidden units, ReLU activations) that outputs four values: height (1), angle (1), and azimuth as a 2D unit vector $(\cos\phi, \sin\phi)$ (2). The architecture is illustrated in Fig.~\ref{fig:architecture}.

For the backbone, we evaluate three architectures: ResNet-50~\cite{resnet} (25.6M parameters), EfficientNetV2-M~\cite{efficientnetv2} (54M), and DINOv3 ConvNeXt-Base~\cite{dinov3} (89M). Each backbone provides four intermediate feature maps to the FPN: layers 1--4 for ResNet-50 and DINOv3 ConvNeXt-Base, and layers 2, 3, 5, 7 for EfficientNetV2-M. As shown in the experiments (Sec.~\ref{sec:experiments}), the DINOv3 backbone---which benefits from self-supervised pretraining via knowledge distillation---substantially outperforms the alternatives.

\subsection{Loss Functions}
\label{sec:loss}

Given an input aerial image $I$, the model learns a function $f_\theta: I \rightarrow \{(b_i, m_i, h_i, \alpha_i, \phi_i)\}_{i=1}^{N}$, where each detected roof segment $i$ has a bounding box $b_i$, binary mask $m_i$, height $h_i$, roof angle $\alpha_i$, and azimuth $\phi_i$. To compute losses, each predicted instance is matched to a unique ground-truth segment via Hungarian matching~\cite{hungarian} on mask IoU; unmatched predictions are treated as background for classification and ignored for attribute regression.

\begin{figure}[t]
  \centering
  \includegraphics[width=\linewidth]{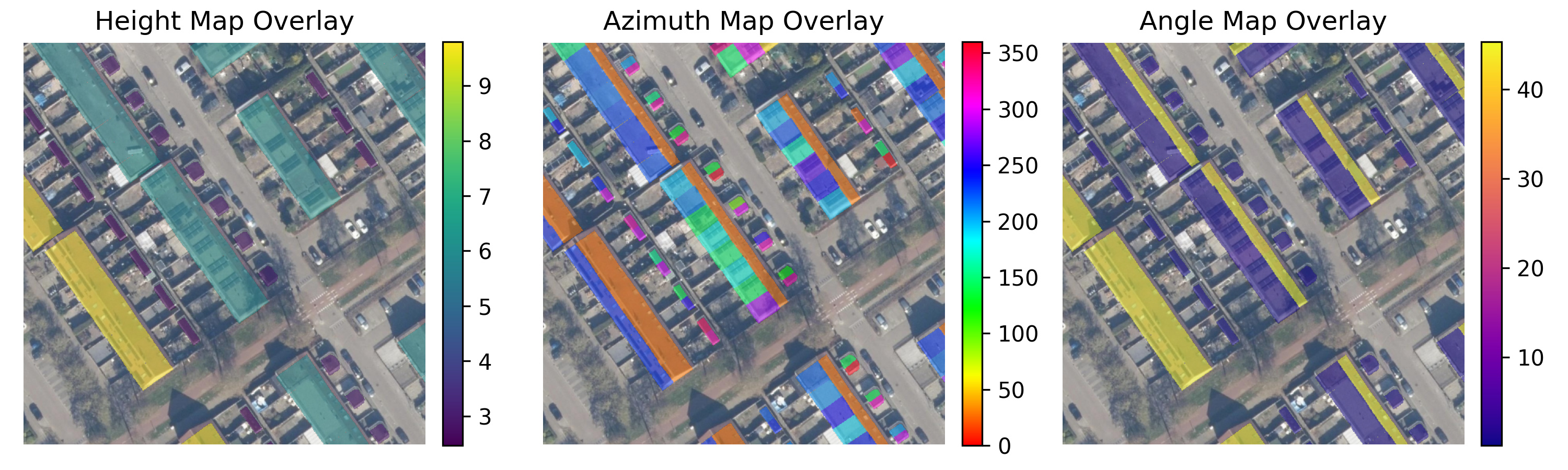}
  \caption{Motivation for the conditional azimuth loss (ground truth data). In the angle map (right), purple segments indicate flat roofs (low slope). These roof segments reveal highly variable colors with no spatial coherence in the azimuth map (center), as horizontal surfaces have no meaningful compass orientation. In contrast, sloped roofs show consistent azimuth patterns.}
  \label{fig:azimuth_motivation}
\end{figure}

The total training loss combines the standard Mask R-CNN losses with an attribute regression term:
\begin{equation}
  \mathcal{L}_{\text{total}} = \mathcal{L}_{\text{RPN}} + \mathcal{L}_{\text{cls}} + \mathcal{L}_{\text{bbox}} + \mathcal{L}_{\text{mask}} + \mathcal{L}_{\text{attr}}.
\end{equation}
The first four terms follow the standard Mask R-CNN formulation. The attribute loss is:
\begin{equation}
  \mathcal{L}_{\text{attr}} = \lambda_h \mathcal{L}_{\text{height}} + \lambda_\alpha \mathcal{L}_{\text{angle}} + \lambda_\phi \mathcal{L}_{\text{azimuth}} \cdot \mathds{1}[\alpha^* > \alpha_{\text{th}}],
\end{equation}
where $\lambda_h = 0.5$, $\lambda_\alpha = 0.001$, $\lambda_\phi = 1.0$, and $\alpha_{\text{th}} = 15\degree$, chosen so the individual loss terms operate on similar scales. Height and angle are supervised with mean squared error:
\begin{equation}
  \mathcal{L}_{\text{height}} = \frac{1}{N}\sum_i (h_i - h_i^*)^2, \quad
  \mathcal{L}_{\text{angle}} = \frac{1}{N}\sum_i (\alpha_i - \alpha_i^*)^2.
\end{equation}
For azimuth, MSE is unsuitable due to the circular discontinuity at $0\degree / 360\degree$. Instead, we use a cosine-based loss (Eq.~\ref{eq:azimuth_loss}). We highlight two design choices:

\paragraph{Conditional azimuth loss.}
Flat roof segments in 3DBAG have arbitrary azimuth labels, since a horizontal surface has no meaningful orientation (Fig.~\ref{fig:azimuth_motivation}). Naively supervising on these labels injects noise into training. We gate the azimuth loss with an indicator function based on the ground-truth roof angle: when $\alpha^* \leq 15\degree$, the azimuth loss is set to zero. To handle the circular nature of azimuth ($0\degree = 360\degree$), we encode the angle as a 2D unit vector $(\cos\phi, \sin\phi)$---a representation also known as sin-cos encoding or \emph{biternion} representation~\cite{biternion}, the 1D analogue of quaternion orientation encoding that has proven effective for continuous angle regression. We supervise with a cosine-based loss:
\begin{equation}
  \mathcal{L}_{\text{azimuth}} = \frac{1}{N}\sum_i \left(1 - \cos(\phi_i - \phi_i^*)\right).
  \label{eq:azimuth_loss}
\end{equation}
This loss reduces to one minus the dot product between the predicted and ground-truth unit vectors, avoiding the need to recover the angle during training. At inference time, the azimuth is decoded from the predicted vector $(c, s)$ via $\phi = \operatorname{atan2}(s, c)$, yielding $\phi \in (-\pi, \pi]$, which is mapped to $[0\degree, 360\degree)$ for evaluation.

\paragraph{Height representation.}
The building height distribution is heavily right-skewed: most buildings are below 20~m, but rare structures (church towers, high-rises) reach ${\sim}100$~m. Direct regression on raw heights causes tall buildings to dominate the gradient signal. We evaluate four alternative compression representations~(Table~\ref{tab:height}):
\begin{align}
  h_i^{\text{linear}} &= h_i / 110, \label{eq:h_linear} \\
  h_i^{\log} &= \log(h_i), \label{eq:h_log} \\
  h_i^{\log_{100}} &= \log_{100}(h_i), \label{eq:h_log100} \\
  h_i^{\text{norm}} &= \frac{\log(h_i) - \mu}{\sigma}, \label{eq:h_lognorm}
\end{align}
where $\mu$ and $\sigma$ are the mean and standard deviation of log-heights computed across all training instances ($\mu = 2.06$, $\sigma = 0.45$). Linear scaling (Eq.~\ref{eq:h_linear}) maps heights to $[0,1]$ but does not address the skew. The logarithmic variants (Eqs.~\ref{eq:h_log}--\ref{eq:h_log100}) compress large values, reducing the influence of rare tall buildings. The log-normalized representation (Eq.~\ref{eq:h_lognorm}) further standardizes to zero mean and unit variance, reshaping the target distribution to be approximately Gaussian and giving the model a balanced gradient signal across all height ranges. At inference, predictions are mapped back to meters by inverting the respective transformation (e.g., de-standardize and exponentiate for log-normalized).

\subsection{Training}
\label{sec:training}

Training follows a two-stage procedure. In the first stage (8 epochs), only the Mask R-CNN components are trained so that the model learns reliable building detections. In the second stage (8 epochs), the backbone is frozen and all four prediction branches---bounding box, classification, mask, and attribute regression---are trained jointly. This staging is essential because the attribute branch relies on correctly detected and matched instances.

We train with Adam using a learning rate of $10^{-3}$ for all parameters except the backbone ($10^{-4}$), a batch size of 16 across 4$\times$A100 GPUs, and float16 mixed precision. We apply dihedral group augmentation (all 8 rotations and reflections of a square) and color jittering (brightness, contrast, and saturation each sampled uniformly from $[0.8, 1.2]$; hue disabled) during training. Images are scaled to $512 \times 512$ resolution for all experiments during training, unless otherwise noted.

\subsection{Evaluation Metrics}
\label{sec:evaluation}

For instance segmentation, we report Average Precision at IoU thresholds of 0.5 (AP$_{50}$), 0.75 (AP$_{75}$), and the COCO mean Average Precision (mAP) averaged over IoU $\in [0.50 : 0.05 : 0.95]$~\cite{coco}. For attribute regression, we report mean absolute error (MAE) computed over predicted instances matched to a ground-truth segment at IoU~$> 0.5$. For height and roof angle, MAE uses the absolute difference directly. For azimuth, the cyclic nature of angles requires the shortest angular distance:
\begin{equation}
  d_{\text{azimuth}}(\phi, \phi^*) = \min\!\left(|\phi - \phi^*|,\; 360\degree - |\phi - \phi^*|\right),
\end{equation}
which replaces the absolute difference in the MAE computation.

\section{Experiments}
\label{sec:experiments}

We evaluate our model through a series of ablation studies, analyzing the impact of backbone architecture, loss design, height representation, and input resolution. We then examine per-cluster performance across different roof types and present qualitative results. All metrics are reported on the held-out test set.

\subsection{Backbone Comparison}
\label{sec:backbone}

Table~\ref{tab:backbone} compares three backbone architectures after the first training stage (8 epochs, detection only). The DINOv3 ConvNeXt-Base backbone substantially outperforms both ResNet-50 and EfficientNetV2-M, achieving an mAP of 0.204 compared to 0.129 and 0.111, respectively. The gap is especially pronounced at AP$_{75}$ (0.177 vs.\ 0.068 and 0.043), indicating that self-supervised pretraining via knowledge distillation produces features that enable more precise mask boundaries. Since accurate instance detection is a prerequisite for attribute regression, all subsequent experiments use DINOv3.

\begin{table}[t]
  \centering
  \small
  \begin{tabular*}{\columnwidth}{@{\extracolsep{\fill}}lccc}
    \toprule
    Backbone & AP$_{50}$ & AP$_{75}$ & mAP \\
    \midrule
    ResNet-50~\cite{resnet}          & 0.337 & 0.068 & 0.129 \\
    EfficientNetV2-M~\cite{efficientnetv2} & 0.320 & 0.043 & 0.111 \\
    DINOv3 ConvNeXt-Base~\cite{dinov3}     & \textbf{0.426} & \textbf{0.177} & \textbf{0.204} \\
    \bottomrule
  \end{tabular*}
  \caption{Instance segmentation performance by backbone (first training stage only). DINOv3 with self-supervised pretraining outperforms both alternatives, particularly at stricter IoU thresholds.}
  \label{tab:backbone}
\end{table}

\subsection{Loss Ablations}
\label{sec:ablations}

\paragraph{Conditional azimuth loss.}
Table~\ref{tab:loss} isolates the effect of different attribute loss configurations. Training all three attributes jointly yields an azimuth MAE of $21.8\degree$, only marginally better than the azimuth-only baseline ($22.3\degree$). Applying the conditional loss---which gates azimuth supervision on $\alpha^* > 15\degree$---reduces azimuth MAE to $12.9\degree$, a 41\% improvement, while angle and height errors remain comparable. This confirms that flat-roof azimuth labels in 3DBAG are essentially noise, and that suppressing them during training allows the model to learn meaningful orientation signals from sloped roofs.

\begin{table}[t]
  \centering
  \small
  \begin{tabular*}{\columnwidth}{@{\extracolsep{\fill}}lccc}
    \toprule
    Loss config.       & Azim. MAE & Angle MAE & Height MAE \\
    \midrule
    Only azimuth       & 22.284\degree & --     & --    \\
    Only angle         & --     & 7.382\degree  & --    \\
    Only height        & --     & --     & \textbf{1.614}\,m \\
    All                & 21.811\degree & 7.388\degree  & 1.704\,m \\
    All + conditional  & \textbf{12.851}\degree & \textbf{7.363}\degree & 1.663\,m \\
    \bottomrule
  \end{tabular*}
  \caption{Attribute regression under different loss configurations. The conditional azimuth loss reduces azimuth error by 41\% without degrading other attributes.}
  \label{tab:loss}
\end{table}

\paragraph{Height representation.}
Table~\ref{tab:height} compares height processing functions. Direct regression on raw heights yields an MAE of 1.878~m. Linear scaling and log-base-100 offer no improvement. Natural log reduces the error to 1.732~m, but the best result (1.556~m, a 17\% improvement over the baseline) comes from the log-normal transformation, which standardizes log-heights to zero mean and unit variance. This reshapes the heavily skewed height distribution into an approximately Gaussian target, providing balanced gradients across all height ranges.

\begin{table}[t]
  \centering
  \small
  \begin{tabular*}{\columnwidth}{@{\extracolsep{\fill}}lc}
    \toprule
    Height representation & Height MAE [m] \\
    \midrule
    None (raw)                        & 1.878 \\
    Linear ($h/110$)                  & 1.966 \\
    $\log$                            & 1.732 \\
    $\log_{100}$                      & 1.963 \\
    $\log$-normalized                 & \textbf{1.556} \\
    \bottomrule
  \end{tabular*}
  \caption{Impact of height representation on regression error. Log-normalized targets yield the lowest MAE by reshaping the skewed distribution into an approximately Gaussian target.}
  \label{tab:height}
\end{table}

\paragraph{Augmentation ablation.}
Table~\ref{tab:augmentation} compares data augmentation strategies using the DINOv3 backbone. Dihedral augmentation (rotations and reflections) provides the largest gain, raising mAP from 0.157 to 0.189. Color jittering alone has a modest effect ($\text{mAP}=0.163$). Combining both yields the best performance ($\text{mAP}=0.204$, AP$_{75}=0.177$), confirming that geometric variation is more critical than photometric variation for roof segmentation, though combining both captures complementary variability.

\begin{table}[t]
  \centering
  \small
  \begin{tabular*}{\columnwidth}{@{\extracolsep{\fill}}lccc}
    \toprule
    Augmentation & AP$_{50}$ & AP$_{75}$ & mAP \\
    \midrule
    None                     & 0.388 & 0.110 & 0.157 \\
    Dihedral                 & 0.403 & 0.147 & 0.189 \\
    Color jitter             & 0.396 & 0.114 & 0.163 \\
    Dihedral + color jitter  & \textbf{0.426} & \textbf{0.177} & \textbf{0.204} \\
    \bottomrule
  \end{tabular*}
  \caption{Effect of data augmentation on instance segmentation (DINOv3 backbone, first training stage). Geometric augmentation dominates; combining both strategies yields the best results.}
  \label{tab:augmentation}
\end{table}

\paragraph{Attribute head capacity.}
Table~\ref{tab:capacity} varies the number of MLP layers and hidden units in the attribute head. A single layer with 64 units is insufficient, yielding high errors across all attributes. Increasing depth and width improves performance up to 2~layers with 128 hidden units, which achieves the best height MAE (1.148~m) and competitive azimuth and angle errors. Larger heads (256 units) offer no further benefit, suggesting that moderate capacity suffices for the attribute regression task. All subsequent experiments use 2~layers with 128 hidden units.

\begin{table}[t]
  \centering
  \small
  \begin{tabular*}{\columnwidth}{@{\extracolsep{\fill}}ccccc}
    \toprule
    Layers & Hidden & Azim. MAE & Angle MAE & Height MAE \\
    \midrule
    1 & 64  & 15.36\degree & 8.43\degree & 2.154\,m \\
    2 & 64  & 12.85\degree & 7.36\degree & 1.663\,m \\
    3 & 64  & \textbf{7.83}\degree & 4.94\degree & 1.309\,m \\
    1 & 128 & 8.71\degree & 5.00\degree & 1.247\,m \\
    2 & 128 & 8.38\degree & \textbf{4.94}\degree & \textbf{1.148}\,m \\
    3 & 128 & 8.34\degree & 5.12\degree & 1.161\,m \\
    1 & 256 & 7.97\degree & 4.97\degree & 1.220\,m \\
    2 & 256 & 8.01\degree & 5.10\degree & 1.324\,m \\
    3 & 256 & 8.69\degree & 5.03\degree & 1.208\,m \\
    \bottomrule
  \end{tabular*}
  \caption{Attribute head capacity study. Varying MLP layers and hidden units with conditional loss. Two layers with 128 hidden units provides the best overall trade-off.}
  \label{tab:capacity}
\end{table}

\subsection{Input Resolution}
\label{sec:resolution}

Increasing input resolution from $512 \times 512$ to $768 \times 768$ improves all metrics (Table~\ref{tab:resolution}). AP$_{50}$ increases from 0.424 to 0.566, mAP from 0.190 to 0.298, and attribute MAEs decrease across the board---azimuth from $8.4\degree$ to $7.0\degree$, angle from $4.9\degree$ to $4.1\degree$, and height from 1.148~m to 1.025~m. This is expected: at $768 \times 768$, each pixel covers approximately 13~cm, preserving fine-grained roof details that are lost at lower resolution.

\begin{table}[t]
  \centering
  \small
  \begin{tabular*}{\columnwidth}{@{\extracolsep{\fill}}lcccccc}
    \toprule
    Res. & AP$_{50}$ & AP$_{75}$ & mAP & Azim. & Angle & Height \\
    \midrule
    $512\!\times\!512$ & 0.424 & 0.144 & 0.190 & 8.38\degree & 4.94\degree & 1.15\,m \\
    $768\!\times\!768$ & \textbf{0.566} & \textbf{0.295} & \textbf{0.298} & \textbf{6.96}\degree & \textbf{4.08}\degree & \textbf{1.03}\,m \\
    \bottomrule
  \end{tabular*}
  \caption{Effect of input resolution on segmentation and attribute regression. Higher resolution consistently improves all metrics.}
  \label{tab:resolution}
\end{table}

\begin{figure}[b]
  \centering
  \includegraphics[width=0.49\linewidth]{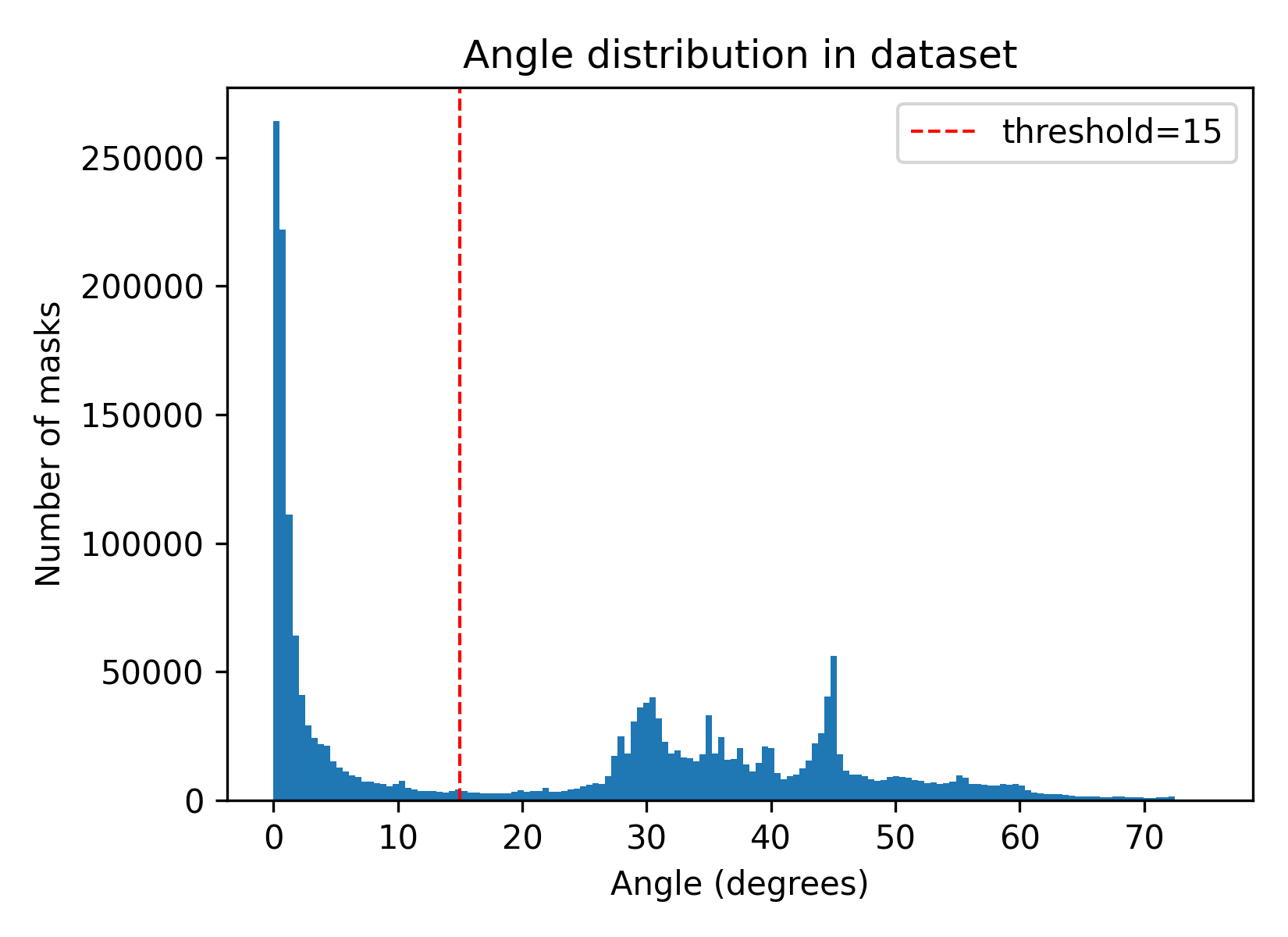}
  \includegraphics[width=0.49\linewidth]{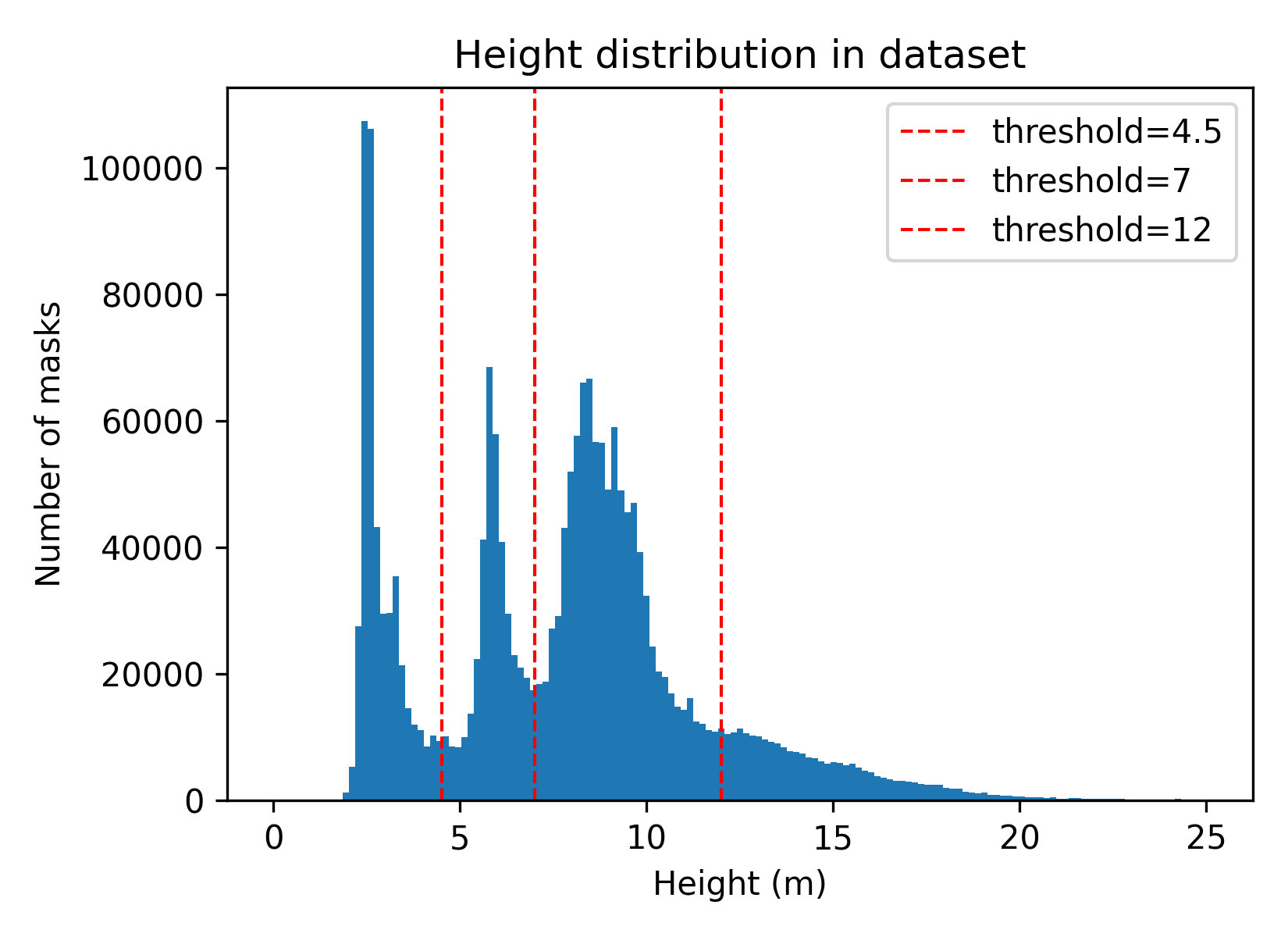}
  \caption{Distribution of roof segment angles (left) and heights (right) across the dataset. Roof angles are divided into flat and steep categories using a $15\degree$ threshold. Heights are divided into four groups: low, medium, high, and very high, using thresholds of 4.5\,m, 7\,m, and 12\,m.}
  \label{fig:cluster_definitions}
\end{figure}

\begin{figure*}[b]
  \centering
  \includegraphics[width=0.9\linewidth]{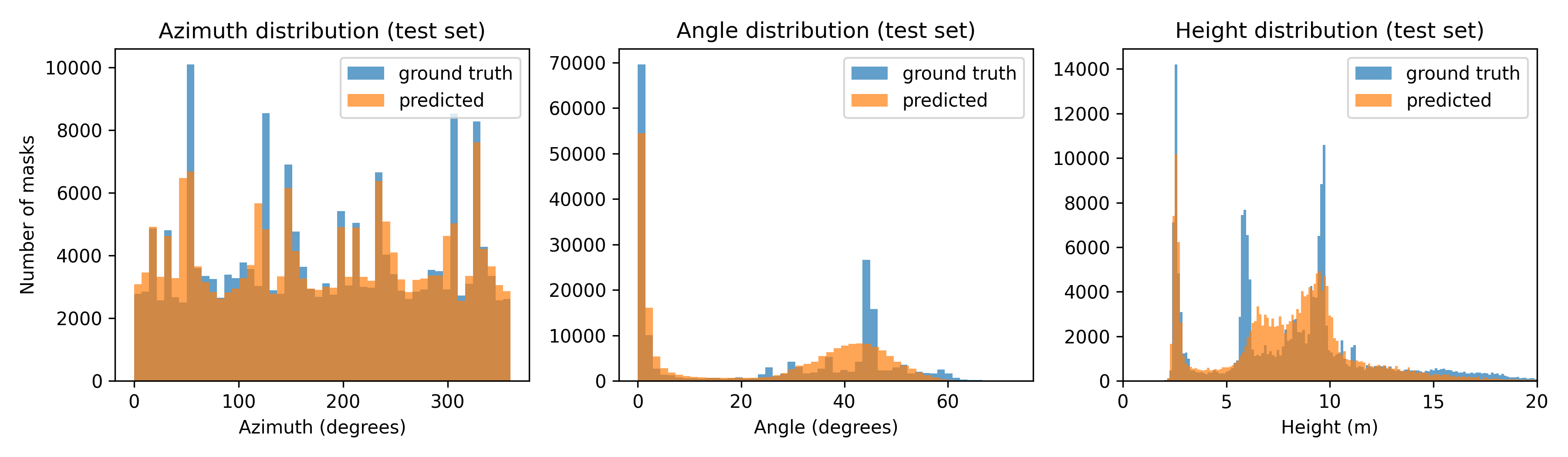}
  \caption{Ground truth (blue) and predicted (orange) distributions for roof segment angle, azimuth, and height on the test set. The predicted distributions show reasonable alignment with the ground truth, indicating that the model captures the overall statistical properties of the geometric attributes.}
  \label{fig:gt_pred_distributions}
\end{figure*}

\subsection{Per-Cluster Analysis}
\label{sec:clusters}

To understand performance variation across roof types, we group test-set instances by angle (flat: $\leq 15\degree$, steep: $> 15\degree$) and height (low: $\leq 4.5$~m, medium: $4.5$--$7$~m, high: $7$--$12$~m, very high: $> 12$~m), as shown in Fig.~\ref{fig:cluster_definitions}. The predicted attribute distributions closely match the ground truth (Fig.~\ref{fig:gt_pred_distributions}), confirming that the model captures the global statistical structure. Table~\ref{tab:clusters} reports matching rates via Hungarian matching at IoU~$\geq 0.5$.

Medium and high steep roofs achieve the highest match rates (69--74\%), as their prominent geometric features provide strong visual cues. Flat roofs are harder: match rates decline from 61\% (low) to 41\% (very high), likely because tall flat-roofed buildings have weaker visual boundaries and more complex surroundings. For attributes, angle prediction is most accurate on flat roofs (MAE $< 4\degree$) and degrades for very high buildings. Azimuth is reliable for steep roofs (MAE $< 2\degree$) across all height categories. Height error grows with building height, from 0.21~m (low flat) to 2.36~m (very high steep), reflecting both the inherent difficulty and the skewed training distribution. The full per-cluster error distributions are provided in the appendix (Fig.~\ref{fig:cluster_errors}).

\begin{table}[b]
  \centering
  \small
  \begin{tabular*}{\columnwidth}{@{\extracolsep{\fill}}llcccc}
    \toprule
    Height & Angle & Match & Angle & Azim. & Height \\
    \midrule
    Low       & Flat  & 60.6\% & 0.58\degree & --   & 0.21\,m \\
    Medium    & Flat  & 64.4\% & 1.47\degree & --   & 0.54\,m \\
    High      & Flat  & 47.8\% & 3.26\degree & --   & 0.47\,m \\
    Very high & Flat  & 40.8\% & 3.76\degree & --   & 2.02\,m \\
    \midrule
    Low       & Steep & 37.0\% & 0.01\degree & 1.57\degree & 0.42\,m \\
    Medium    & Steep & 69.3\% & 0.33\degree & 0.79\degree & 0.87\,m \\
    High      & Steep & 74.1\% & 1.68\degree & 0.53\degree & 0.13\,m \\
    Very high & Steep & 39.9\% & 2.99\degree & 0.58\degree & 2.36\,m \\
    \bottomrule
  \end{tabular*}
  \caption{Per-cluster matching rate and attribute errors. Azimuth is reported only for steep roofs, where the ground-truth orientation is meaningful. Medium and high steep roofs are easiest to detect; very high buildings are hardest across all metrics.}
  \label{tab:clusters}
\end{table}

\subsection{Qualitative Results}
\label{sec:qualitative}

Errors tend to concentrate on very tall or geometrically complex buildings, and at boundaries between adjacent roof segments---consistent with the quantitative per-cluster analysis. We illustrate typical success cases and failure modes per attribute below.

\begin{figure*}[t]
  \centering
  \includegraphics[width=0.52\linewidth]{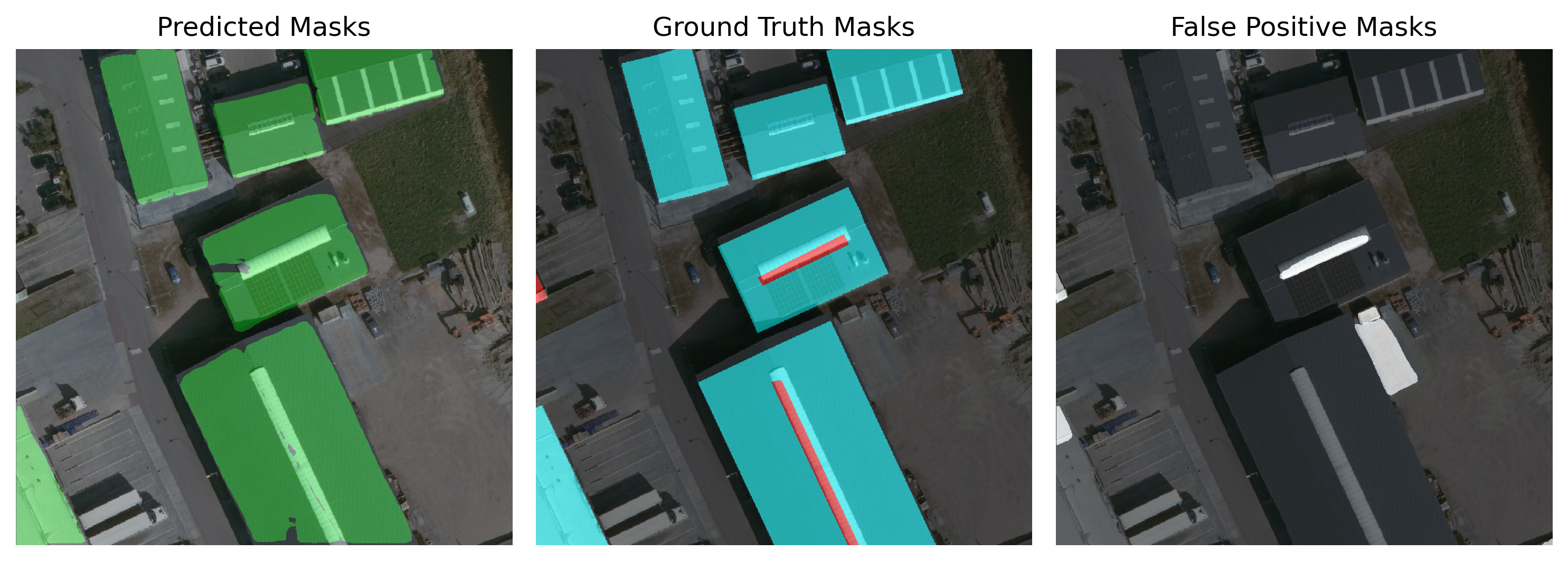}\\[1mm]
  {\small (a) Residential area. The model correctly detects most roof segments.}\\[4mm]
  \vspace{-3mm}
  \includegraphics[width=0.52\linewidth]{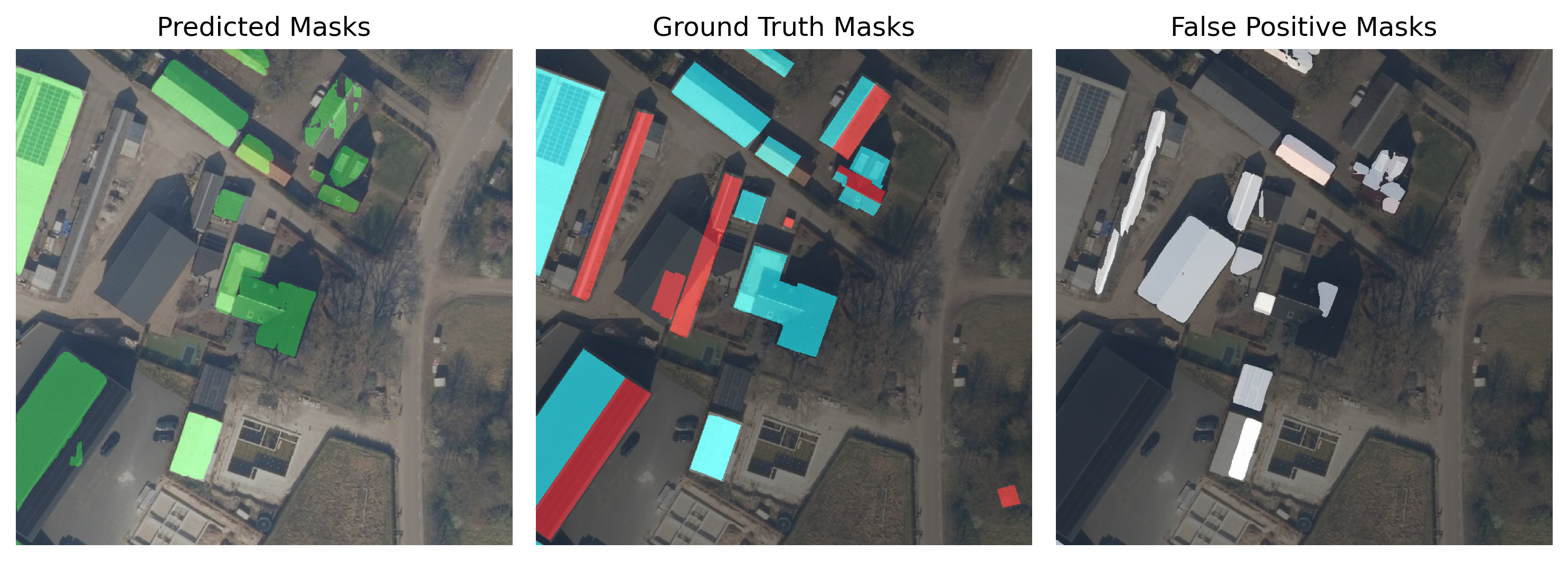}\\[1mm]
  {\small (b) Temporal mismatch: a former barn has been replaced by a house, yet the model correctly segments the current building.}\\[4mm]
  \vspace{-3mm}
  \includegraphics[width=0.52\linewidth]{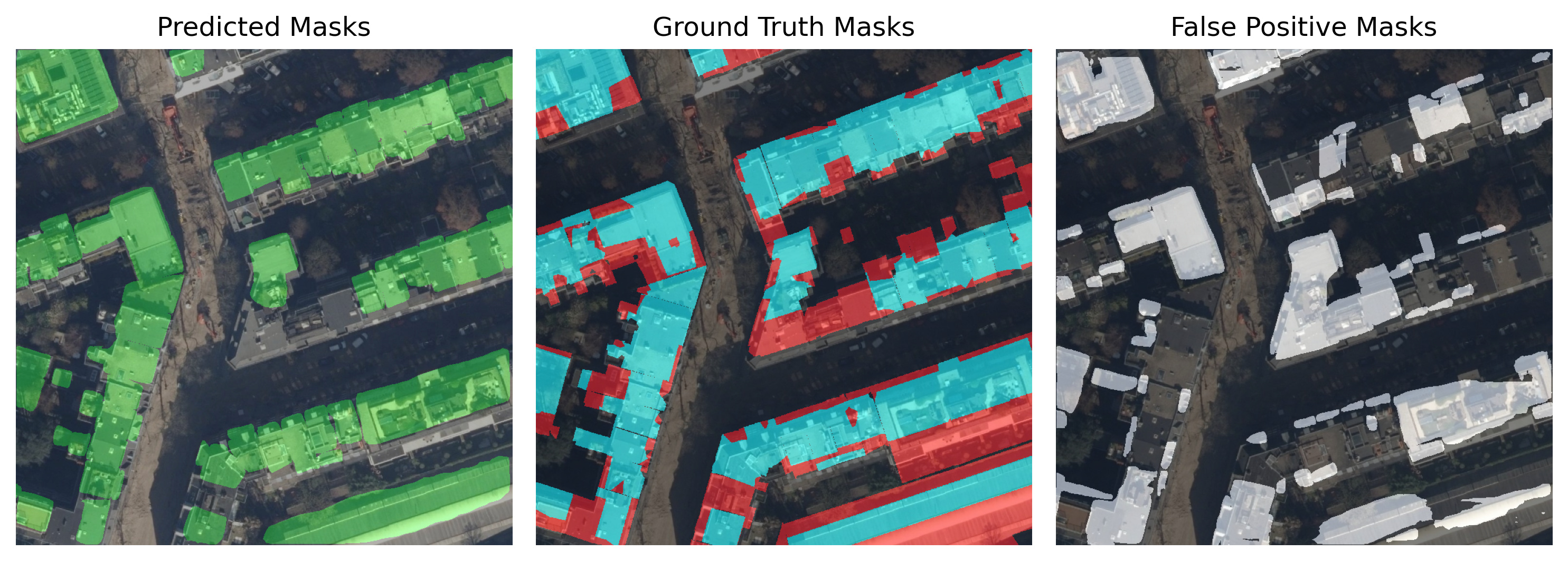}\\[1mm]
  {\small (c) Dense urban area. Most roof segments are detected, but 3DBAG over-segmentation causes missed matches.}
  \caption{Instance segmentation results. Left: predicted masks (green). Middle: matched ground-truth masks (light blue) and missed ground-truth segments (red). Right: false-positive predictions (IoU $< 0.5$). (a) Strong performance on a residential area. (b) The ground truth is outdated and the model prediction is correct. (c) Dense urban scene with complex roof layouts.}
  \label{fig:segmentation}
\end{figure*}

\paragraph{Segmentation.}
Figure~\ref{fig:segmentation} shows the instance segmentation output for a dense urban area with matched, missed, and false-positive masks. The model detects most roofs, but mismatches often stem from annotation granularity rather than detection failures: 3DBAG splits visually coherent surfaces into multiple segments, causing correct predictions to match only one ground-truth part while the rest are counted as false negatives. We also observe cases where the model produces visually plausible predictions that are penalized due to temporal misalignment between the aerial imagery and the 3DBAG annotations (Fig.~\ref{fig:segmentation}b). These observations are discussed further in Sec.~\ref{sec:discussion}.

\begin{figure*}[t]
  \centering
  \includegraphics[width=0.72\linewidth]{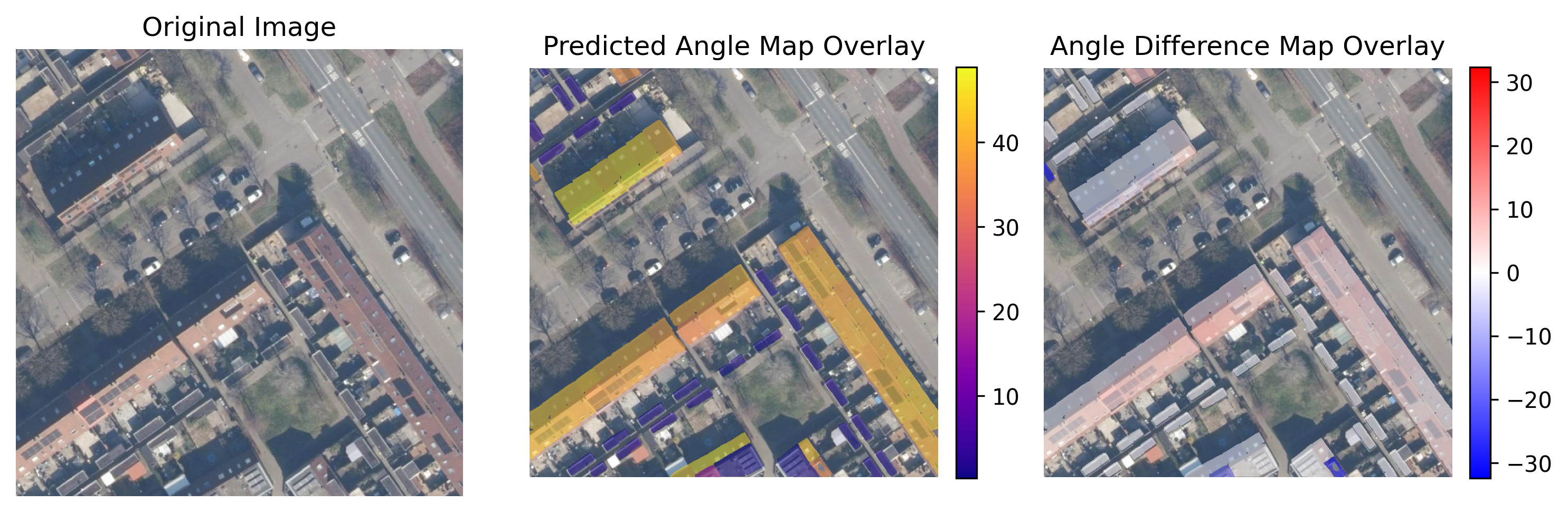}\\[1mm]
  {\small (a) Row houses: predicted angles closely match ground truth.}\\[4mm]
  \vspace{-3mm}
  \includegraphics[width=0.72\linewidth]{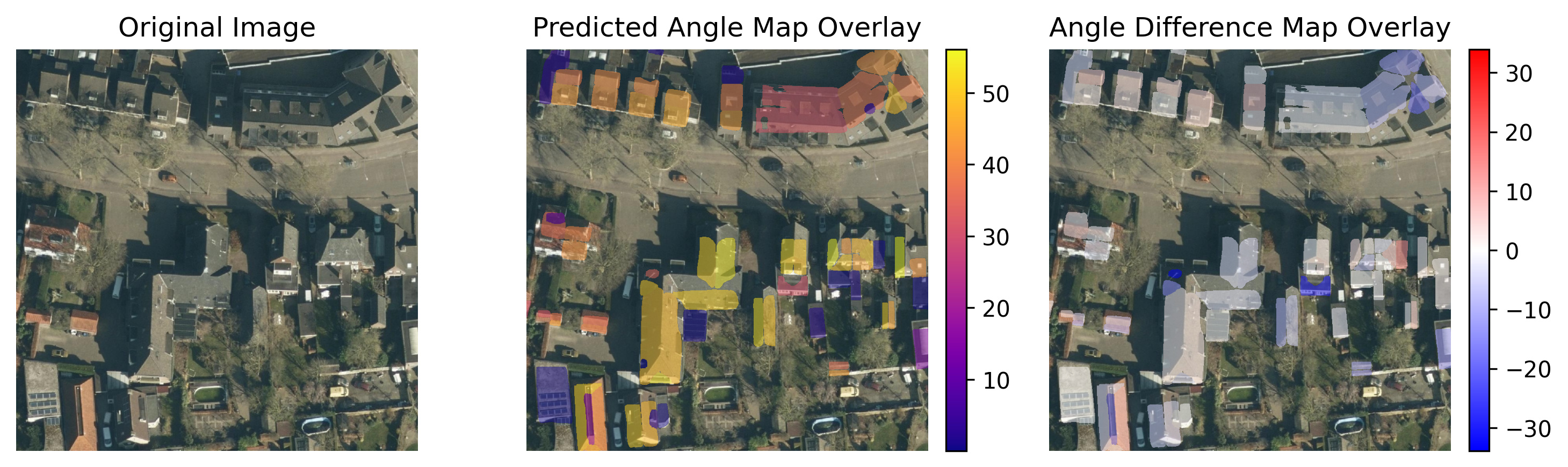}\\[1mm]
  {\small (b) Irregular roofs: accurate estimates, though shadows introduce local errors.}\\[4mm]
  \vspace{-3mm}
  \includegraphics[width=0.72\linewidth]{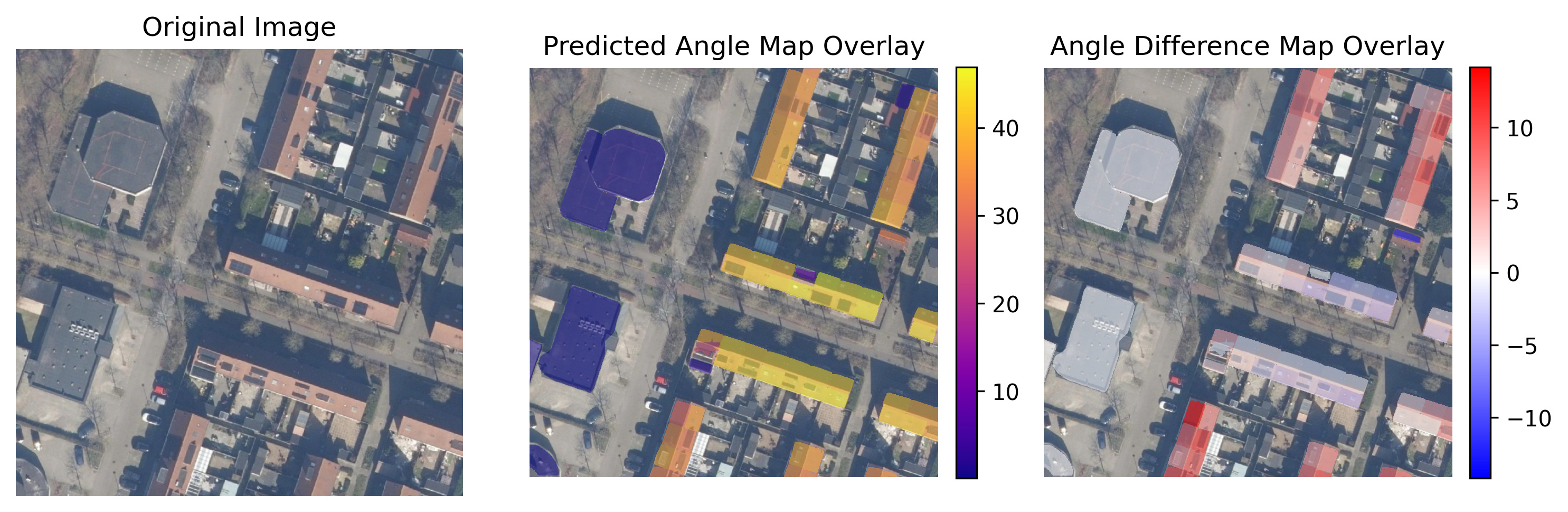}\\[1mm]
  {\small (c) Inconsistent angle estimates across roofs sharing the same true orientation.}
  \caption{Roof angle prediction examples. (a) Well-performing case on structured row houses with small boundary deviations. (b) The model handles irregular roof structures, though shadows introduce localized errors. (c) Failure case: inconsistent angle estimates reflecting the model's reliance on local appearance.}
  \label{fig:angle_qualitative}
\end{figure*}

\paragraph{Roof angle.}
The model predicts roof angles accurately on structured row houses, where strong geometric cues yield consistent estimates with small boundary deviations (Fig.~\ref{fig:angle_qualitative}a). It also generalizes to irregular roof structures, though shadows introduce localized errors where edge information is weak (Fig.~\ref{fig:angle_qualitative}b). A notable failure mode is inconsistency across neighboring roofs that share the same true orientation: subtle image variations such as shadows, texture changes, or slight annotation misalignment cause the model to assign different angle values to adjacent segments (Fig.~\ref{fig:angle_qualitative}c), reflecting its reliance on local appearance without explicit geometric constraints.

\begin{figure*}[t]
  \centering
  \includegraphics[width=0.8\linewidth]{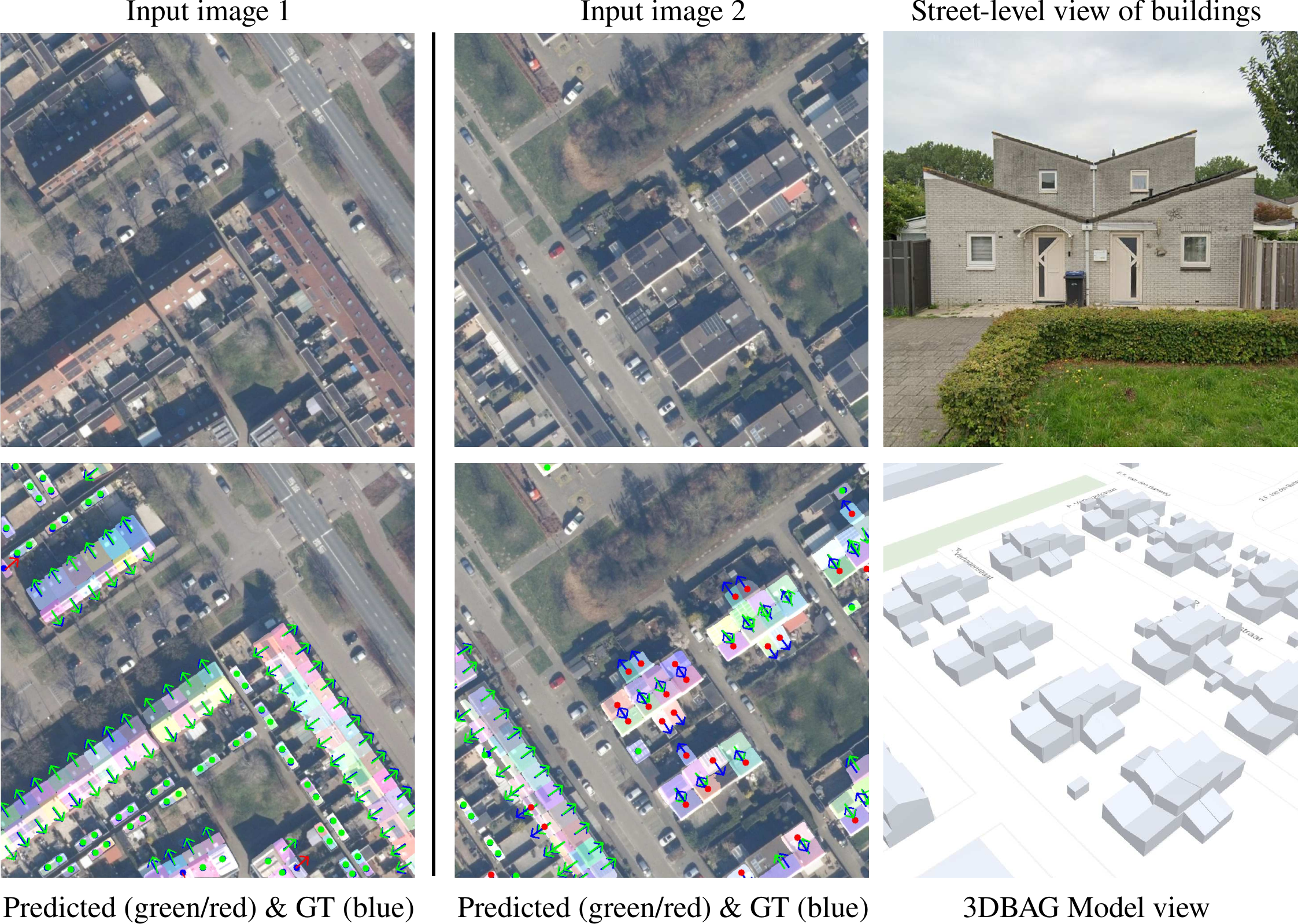}
  \caption{Azimuth prediction examples. (a) The model performs well on structured row houses with consistent orientations. (b) Failure case: inward-sloping roofs are visually ambiguous from a single overhead view, causing the model to predict the opposite azimuth direction.}
  \label{fig:azimuth_qualitative}
\end{figure*}

\paragraph{Azimuth.}
On uniformly oriented row houses with clear ridge lines, the model produces consistent and accurate azimuth estimates (Fig.~\ref{fig:azimuth_qualitative}a). A challenging failure case involves inward-sloping roofs, where both planes tilt toward a central valley (Fig.~\ref{fig:azimuth_qualitative}b). From a nadir viewpoint, this geometry is visually ambiguous---the model frequently predicts these roofs as flat or assigns the opposite azimuth direction. The 3DBAG visualization and street-level imagery confirm the unusual geometry, indicating that the errors reflect genuine input ambiguity rather than a learning failure. Multi-view input would likely help disambiguate such cases.

\begin{figure*}[t]
  \centering
  \begin{minipage}[t]{0.48\linewidth}
    \centering
    \includegraphics[width=\linewidth]{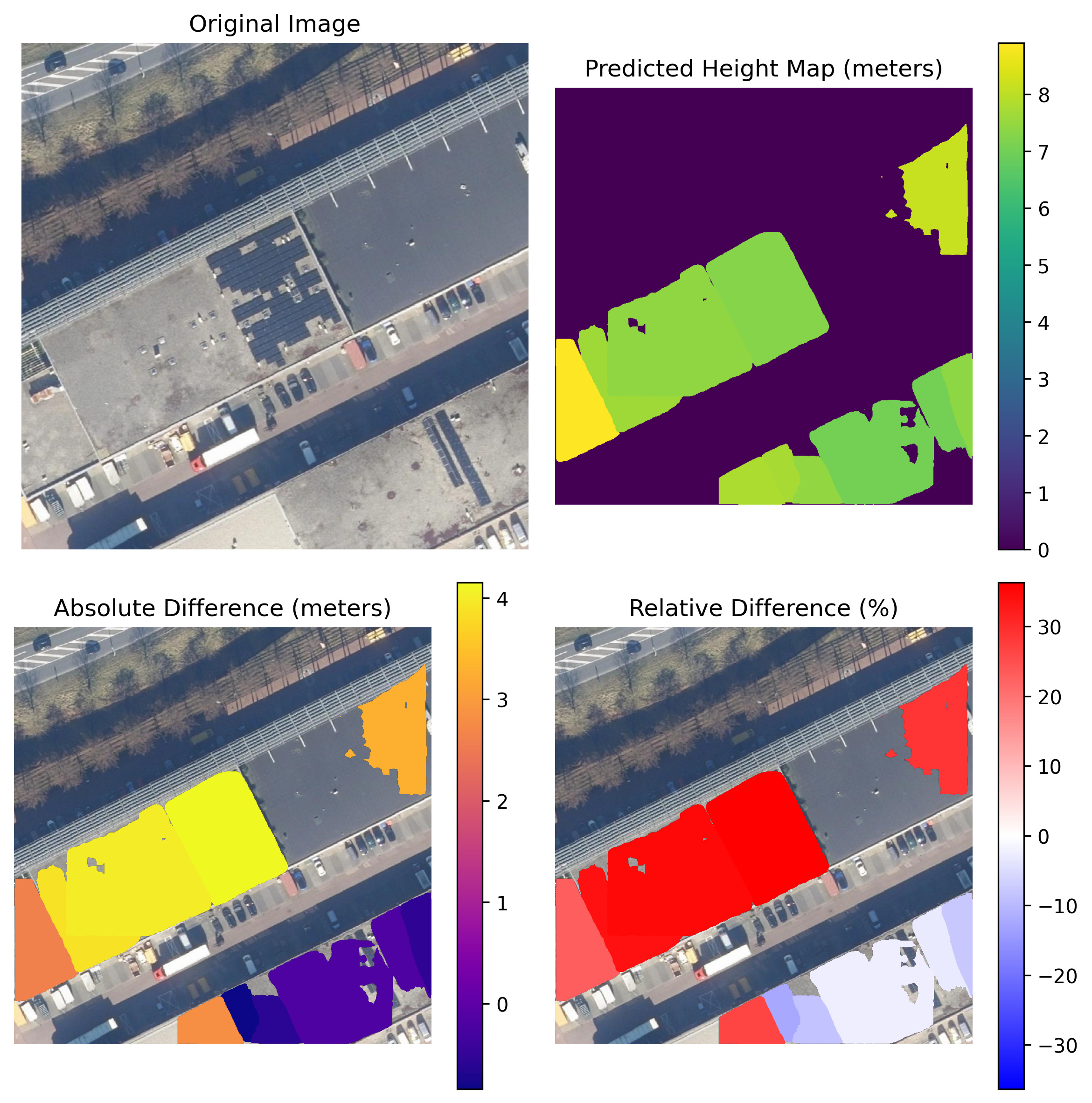}\\[1mm]
    {\small (a) Predicted (top) and ground-truth (bottom) height maps. The model captures the overall height structure, with errors on tall or complex buildings.}
  \end{minipage}\hfill
  \begin{minipage}[t]{0.48\linewidth}
    \centering
    \includegraphics[width=\linewidth]{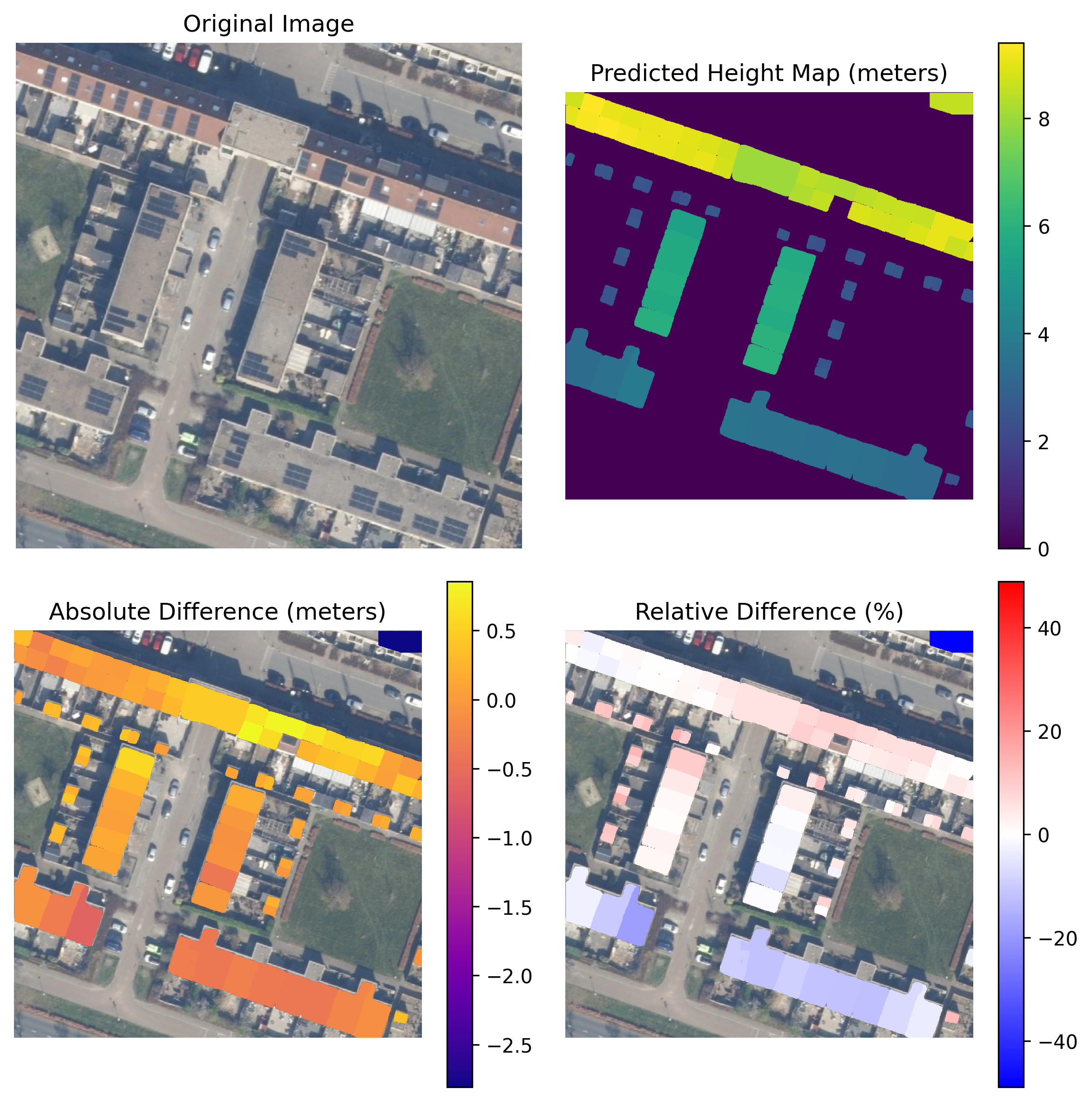}\\[1mm]
    {\small (b) Inconsistent height predictions across neighboring roofs with similar true heights.}
  \end{minipage}
  \caption{Height prediction examples. (a) The model correctly distinguishes low residential buildings from taller structures. (b) Adjacent buildings with similar true heights receive inconsistent predictions, reflecting the model's reliance on local appearance without explicit geometric constraints linking neighboring segments.}
  \label{fig:height_qualitative}
\end{figure*}

\paragraph{Height.}
Height predictions are generally accurate for low and medium buildings (Fig.~\ref{fig:height_qualitative}a) but lack spatial consistency for adjacent structures with similar true heights (Fig.~\ref{fig:height_qualitative}b). The model alternates between under- and overestimation across neighboring segments, reflecting its treatment of each instance independently without explicit geometric constraints linking adjacent roof segments.

\section{Discussion}
\label{sec:discussion}

\paragraph{Strengths and failure modes.}
The per-cluster analysis reveals clear patterns in where the model succeeds and fails. Medium and high steep roofs form the strongest category, with match rates above 69\% and small errors across all three attributes, as their prominent edges and shading provide strong visual cues. Low flat roofs are also predicted reliably: slope errors are near zero and height errors remain below 0.55~m. Performance degrades for low-steep roofs, where weak shading and subtle slope changes provide insufficient visual signal, and for very tall buildings, where complex surroundings, stronger shadow effects, and underrepresentation in the training distribution increase errors. We also observe that the model does not enforce spatial coherence between adjacent roof segments: neighboring segments with similar true heights or slopes occasionally receive noticeably different predictions, reflecting the instance-based formulation's reliance on local appearance without explicit geometric constraints linking adjacent segments.

\paragraph{3DBAG as ground truth.}
While 3DBAG is an invaluable ground-truth source---consistent, nationwide, and derived from accurate LiDAR measurements---its characteristics directly affect the reported metrics. A key issue is segmentation granularity: 3DBAG frequently splits a visually coherent roof surface into multiple polygons around ridges, dormers, or minor height differences. Under one-to-one Hungarian matching at IoU~$\geq 0.5$, a single correct prediction covering multiple ground-truth segments can only match one, causing the rest to be counted as false negatives. Similarly, temporal misalignment between the aerial imagery and 3DBAG (newly constructed or demolished buildings) and minor spatial offsets between the two data sources reduce IoU scores for otherwise correct predictions. As a result, qualitative inspection suggests that the model performs better than the quantitative metrics indicate. For example, a roof with solar panels may be split into multiple ground-truth segments, while the model predicts a single coherent mask that is penalized under one-to-one matching.

\paragraph{Future directions.}
Several directions could address the identified limitations. Pre-processing 3DBAG polygons into larger, semantically coherent roof surfaces, or adopting many-to-one evaluation protocols, would align the ground truth more closely with the visual evidence and reduce evaluation artifacts. Expanding and diversifying the training data beyond the Netherlands would improve generalization to other building styles. Incorporating multi-view information, such as oblique aerial or street-level imagery, could help disambiguate complex roof shapes and improve height consistency. Finally, moving beyond purely instance-based prediction to jointly model neighboring segments, or incorporating explicit geometric constraints during training, could encourage spatial coherence and improve suitability for downstream 3D reconstruction.

\section{Conclusion}
\label{sec:conclusion}

This work presented a method for joint instance segmentation and geometric attribute regression of roof structures from single aerial images. The proposed approach extends Mask R-CNN with a dedicated attribute branch that predicts building height, roof slope, and roof azimuth per detected roof segment. Two contributions proved particularly effective: a conditional azimuth loss that suppresses noisy supervision for flat roof segments, reducing azimuth error by 41\%, and a log-normalized height representation that reshapes the skewed target distribution, improving height regression by~17\%.

Trained on 17,353 Dutch aerial images with automatically derived 3DBAG ground truth, the final model achieves mean absolute errors of approximately 7\degree{} for azimuth, 4\degree{} for roof slope, and 1~meter for building height. Instance segmentation reaches an AP$_{50}$ of 0.566. Importantly, expensive 3D reference data is required only during training. At inference time, only a single aerial orthophoto is used.

These results demonstrate that automated geometric building analysis from aerial imagery is feasible at national scale. The method is deployable wherever aerial imagery is available, providing a scalable foundation for applications in energy planning, insurance risk assessment, and sustainable building management.

{
    \small
    \bibliographystyle{ieeenat_fullname}
    \bibliography{main}
}

\appendix

\begin{figure*}[t]
  \centering
  \includegraphics[width=0.85\linewidth]{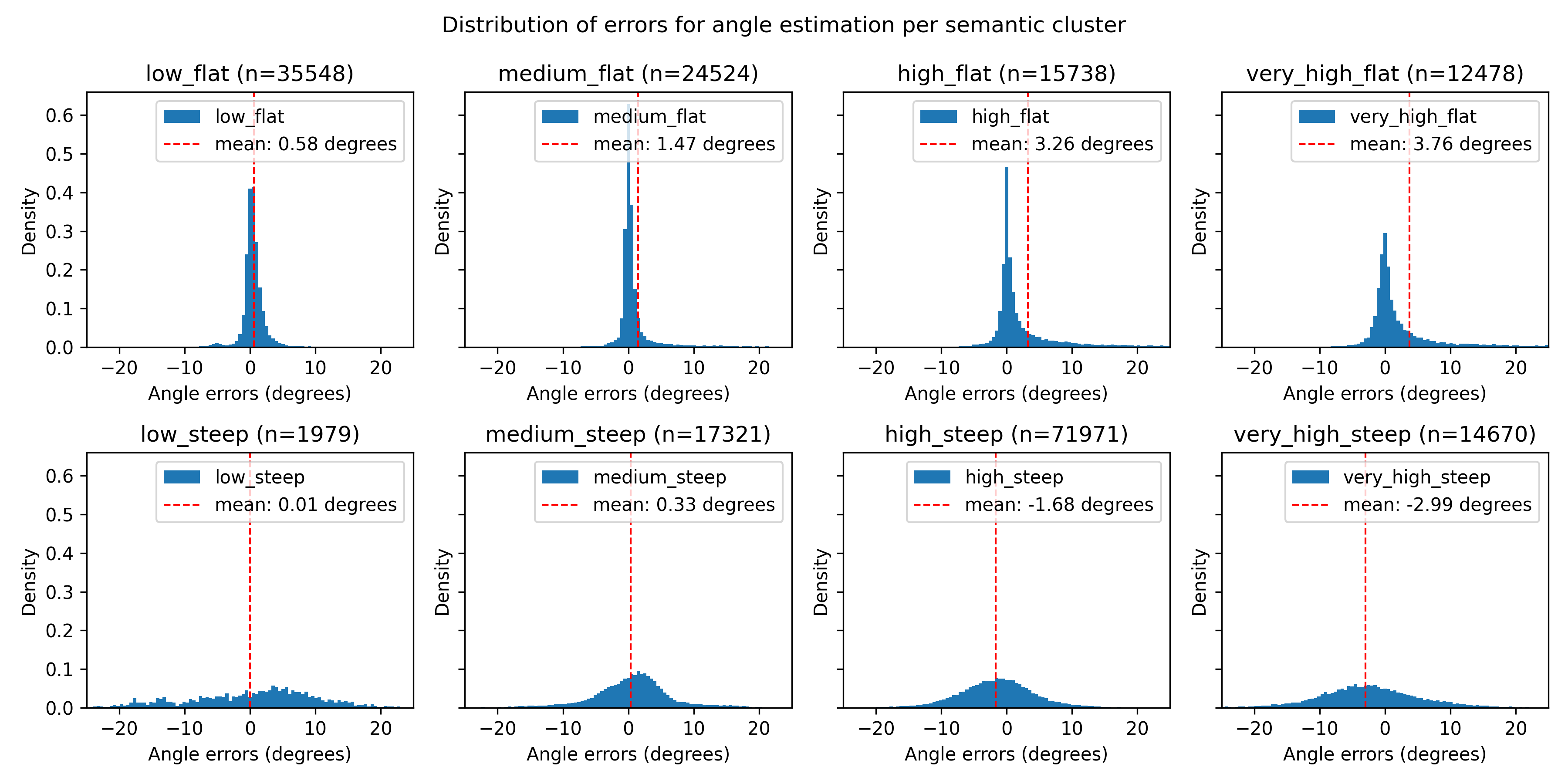}\\[1mm]
  {\small (a) Angle prediction errors per cluster. Top row: flat roofs; bottom row: steep roofs; left to right: low to very high.}\\[4mm]
  \includegraphics[width=0.85\linewidth]{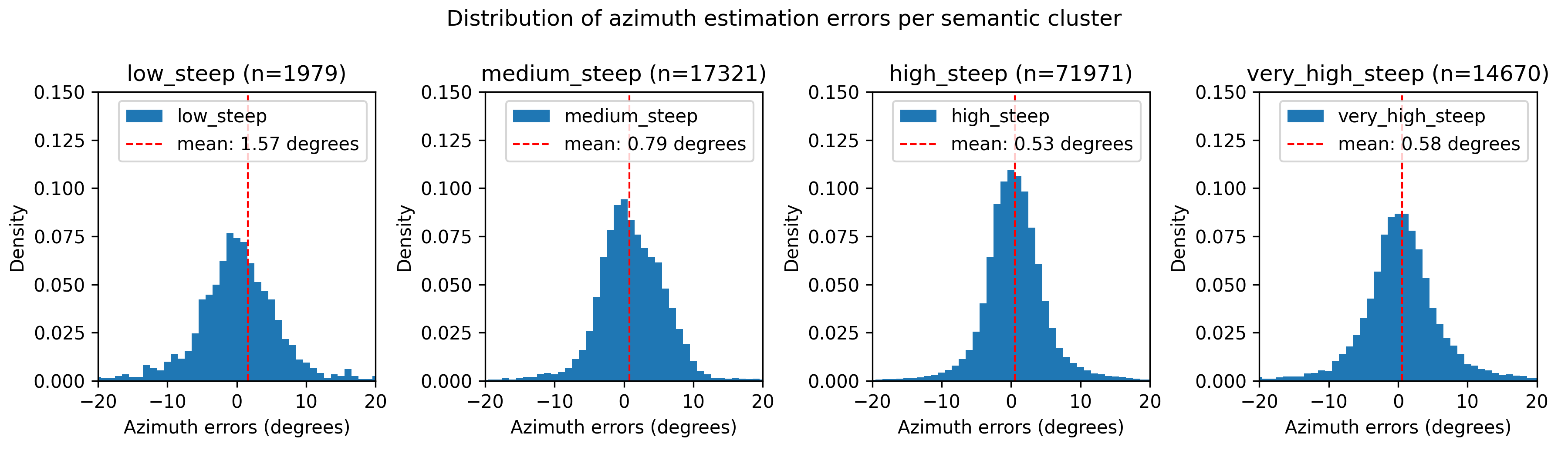}\\[1mm]
  {\small (b) Azimuth prediction errors for steep roofs. Distributions are centered near zero, with most predictions within $\pm 10\degree$.}\\[4mm]
  \includegraphics[width=0.85\linewidth]{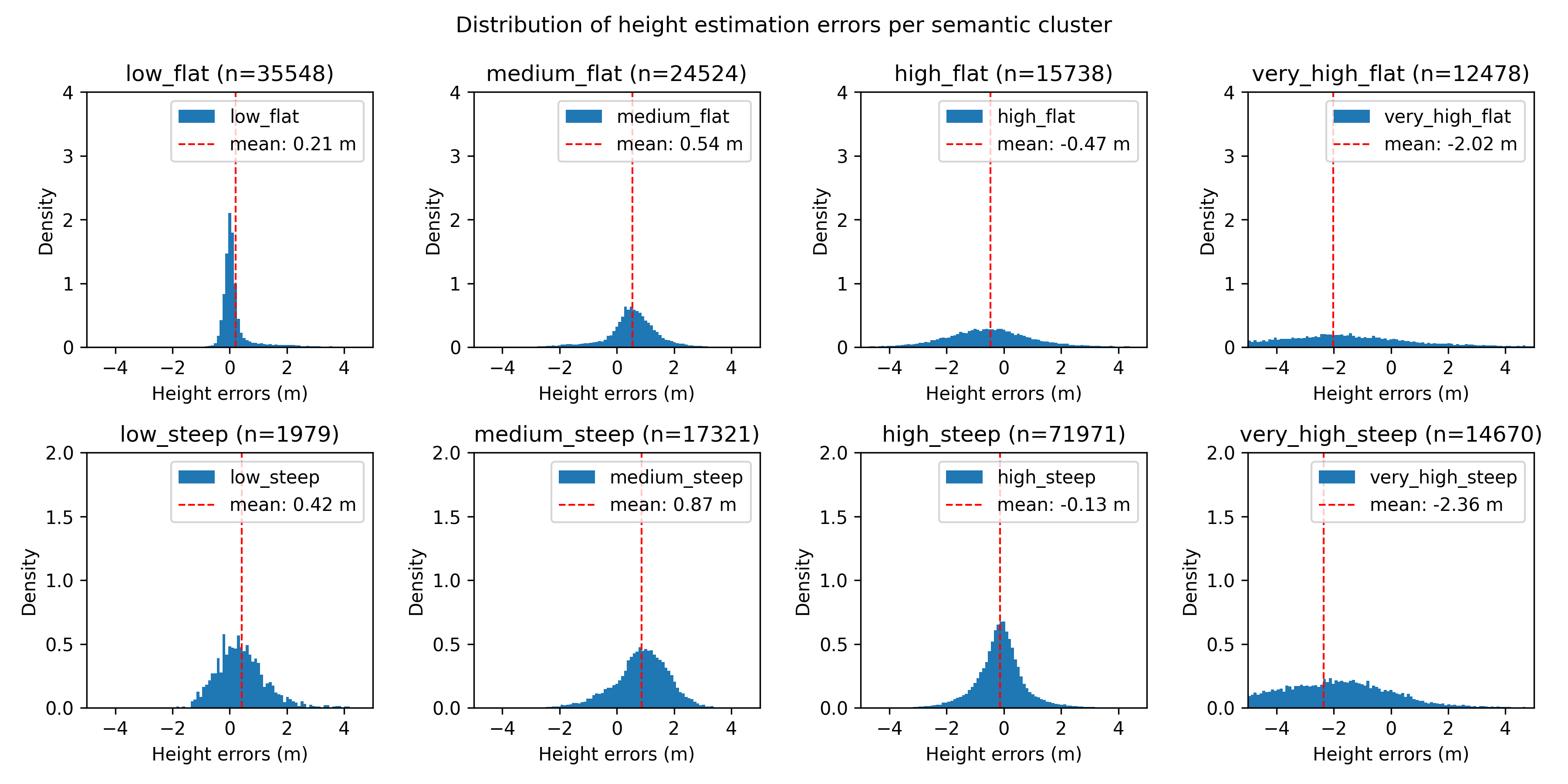}\\[1mm]
  {\small (c) Height prediction errors. Low-height segments show minimal spread; very high segments show the largest errors.}
  \caption{Per-cluster error distributions for (a) roof angle, (b) azimuth, and (c) height. Each row of subplots groups roof segments by height (low to very high) and angle (flat vs.\ steep). Medium and high steep roofs achieve the tightest distributions, while very high buildings show the broadest error spread across all attributes.}
  \label{fig:cluster_errors}
\end{figure*}

\end{document}